\documentclass{article}

\usepackage{graphicx}
\usepackage{xcolor}
\usepackage{subcaption}
\usepackage{booktabs}
\usepackage{amsmath}
\usepackage{amssymb}
\usepackage[inline]{enumitem}

\usepackage[preprint]{corl_2026} 

\title{Path Planning in Physically Viable World Models}

\author{%
\textbf{Su Ann Low} \quad
\textbf{Cheng-Hsi Hsiao} \quad
\textbf{Xingjian Li} \\
\textbf{Adam J. Thorpe} \quad
\textbf{Ufuk Topcu} \quad
\textbf{Krishna Kumar} \\
\normalfont The University of Texas at Austin \\
\normalfont\texttt{suann@utexas.edu} \quad
\normalfont\texttt{chhsiao@utexas.edu} \quad
\normalfont\texttt{xingjian.li@austin.utexas.edu} \\
\normalfont\texttt{adam.thorpe@austin.utexas.edu} \quad
\normalfont\texttt{utopcu@utexas.edu} \quad
\normalfont\texttt{krishnak@utexas.edu}
}

\begin{document}
\maketitle


\begin{abstract}
Robots deployed in unstructured outdoor environments often plan from scene reconstructions collected before deployment because operators cannot remap large or remote sites before every mission. As a result, robots must make long-horizon planning decisions using stale maps that assume the terrain remains unchanged, even though physical changes to the environment may render previously feasible routes unsafe or unreachable at execution time. We present a physically viable world model for evaluating what-if queries for robot navigation under future terrain change. The system augments reconstructed 3D Gaussian splat scenes with physics-based simulation to generate physically modified versions of the same environment without recollecting sensor data or rebuilding the map. We then implement a terrain-aware planner that accounts for physical events, obstacles, and deformations that are simulated by the world model. This allows robots and human operators to evaluate whether planned routes remain feasible before committing to a planned route, particularly in constrained environments where retreat or recovery may become impossible once conditions change. We evaluate the system on a real outdoor field site in Central Texas using simulated flooding across multiple severity levels. We measure route and mission feasibility as terrain conditions deteriorate under physically simulated interventions. Our results show that physically viable world models expose long-horizon route failures and rerouting behavior that are not apparent when planning only on the original reconstructed environment, allowing robots to evaluate how future terrain changes may affect route feasibility before deployment.
\end{abstract}



\section{Introduction}

Most path planning algorithms reason over a single, fixed representation of the environment. Given a map, the planner searches for a feasible or optimal route and assumes that the represented world remains unchanged throughout the mission. This limits planning to asking which path is best in the observed environment, rather than whether the mission remains feasible under a specified physical change. That distinction is critical in off-road autonomy, disaster response, and field robotics, where flooding, terrain collapse, and debris accumulation can alter traversability after a map has been built. In constrained terrain, the cost of this limitation is severe: once a robot commits to a crossing, discovering during execution that it has flooded may leave no safe way back. Planning in such settings therefore requires more than a static map---it requires a physically viable world model (PVWM) that supports query-driven reasoning over families of physically modified scenes \cite{thorpe2026physicallyviable}.

\begin{figure}[!t]
    \centering
    \includegraphics[width=\linewidth, height=2in]{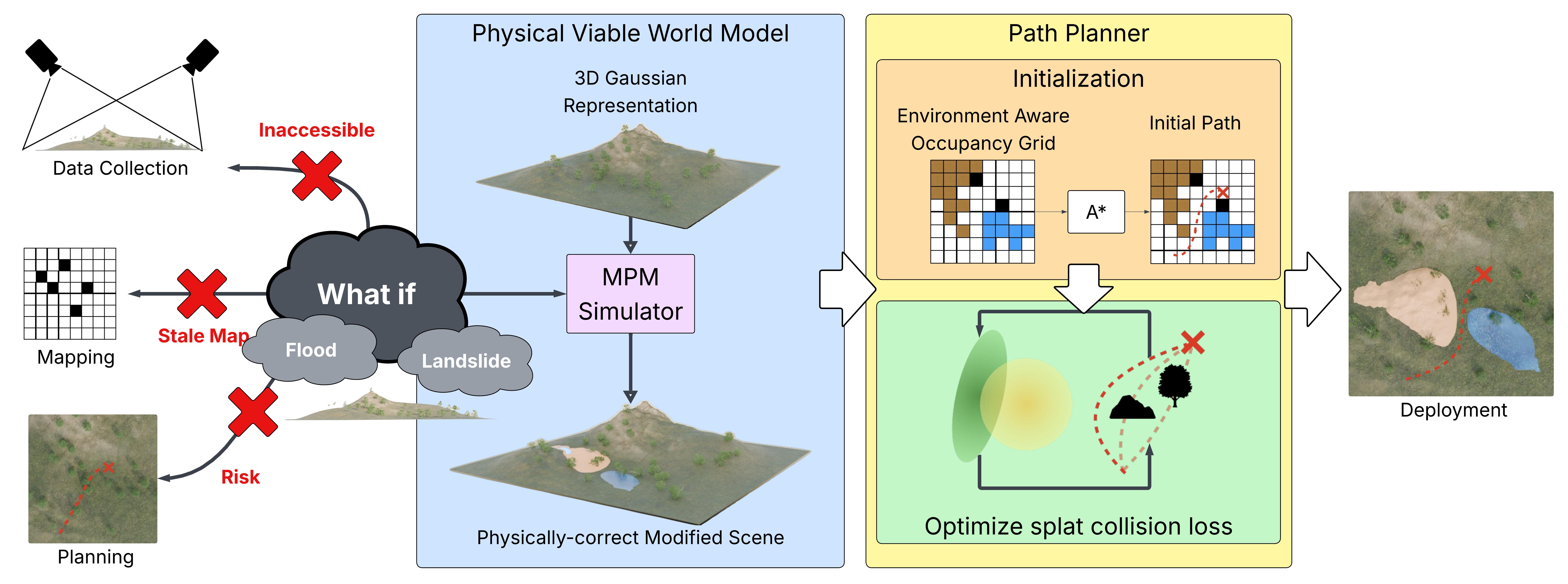}
    \caption{A physically viable world model generates scenes given different queries, and a traversability-aware path planner optimizes for routes under physical viability constraints. }
    \label{fig:pipeline}
\end{figure}

We present a method for path planning for off-road autonomous robots in physically viable world models. A physically viable world model transforms a reconstructed scene into query-conditioned environments generated by specified physical interventions. Such models answer what \emph{would} happen under a given intervention, allowing a robot to evaluate conditions that have not yet occurred. We instantiate this framework using Gaussian splat reconstructions built from historical observations, since recollecting data before each mission is often impractical in large, remote, or time-sensitive environments. Reasoning about a hypothesized event therefore consists of modifying an existing reconstruction and planning in the resulting environment.
We generate these modifications using Material Point Method (MPM) simulation, which models interventions such as flooding and terrain deformation. Planning operates on the modified environments rather than only on the original reconstruction. Because a free-space optimizer may route a path through thin hazard layers that a ground vehicle cannot traverse, our planner constrains paths to the terrain surface, blocks intervention-affected regions according to physical severity, and enforces ground-contact constraints, producing physically realizable routes \cite{andreu2025foci}. A single reconstruction can therefore support planning across many physically generated environments, revealing route failures, rerouting behavior, and reachability loss that are invisible in the observed scene alone.

Our contribution is to formulate path planning as a query over a physically viable world model: from a single reconstructed scene, we evaluate whether a route remains feasible under a specified physical change before the robot commits to it. We evaluate the approach in three realistic scenarios. In a Central Texas field site, increasing flood severity causes routes to transition from persistent paths to topologically distinct detours and ultimately to an unreachable goal. In an Alaska village scene, we add floodwater while keeping the reconstructed geometry fixed to isolate the effect of changing vehicle mobility on route feasibility. In a sandbox scene, we simulate landslides that introduce previously unobserved hazards and force rerouting around the resulting runout. Together, these examples demonstrate how physically modified world models expose route failures, rerouting, and reachability loss that cannot be identified from the observed reconstruction alone.

\paragraph{Contributions.}
We make the following contributions: 
\begin{enumerate*}[label=(\arabic*)]
    \item 
    We construct a physically viable world model that augments an existing Gaussian splat reconstruction using physics-based simulation to produce hypothetical future environmental states for pre-execution route evaluation, without requiring new data collection or scene retraining.
    \item 
    We construct a terrain-following planning pipeline over physically augmented Gaussian splat scenes that enforces ground-contact constraints throughout search and optimization and classifies intervention-affected cells based on intervention severity, enabling pre-execution hazard avoidance in unstructured outdoor environments.
    \item 
    We demonstrate the approach on a real flood-prone, rugged outdoor field site using drone-reconstructed Gaussian splat scenes and physics-based flood simulation, showing how future environmental changes alter feasible routes and long-horizon planning decisions across varying intervention severity.
\end{enumerate*}

\section{Related Work}

\textbf{Off-road autonomy in unstructured environments. }
Autonomous navigation in unstructured environments has largely focused on traversability estimation, terrain classification, local planning, and adaptive control. Terrain-aware cost maps, traversability analysis, and geometric reasoning identify feasible paths and support obstacle avoidance \cite{10684236, chen2024learning-on-the-drive, 10160856, Meng-RSS-23}, while terrain-aware dynamics estimation and online adaptation improve rollout accuracy for model predictive control and trajectory optimization \cite{6859073,kumar2021rma, ward2025onlineadaptationterrainawaredynamics, ward2026zeroautonomyrealtimeonline, doi:10.2514/1.G001921, 10.1109/TRO.2018.2865891, pmlr-v70-finn17a, clavera2018learning}. These approaches improve planning and control under changing conditions, but generally treat the environment as fixed aside from local traversability variation. In contrast, terrain composition, deformation, fluid interaction, vegetation response, and contact mechanics may evolve under interaction and remain only partially observable. Classical planning and robust control methods address uncertainty through receding-horizon optimization, belief-space planning, and uncertainty-aware trajectory generation, but typically assume that the relevant environment representation is already known. In the settings we consider, robots instead operate using incomplete or inaccurate prior maps where the physical properties governing future traversability may be unknown, spatially varying, and interaction-dependent.

\textbf{Planning in Gaussian splats. }
Neural scene representations enable planning directly within learned environment models. NeRF-based methods use learned density fields for trajectory optimization and chance-constrained planning \cite{pantic2022sampling, adamkiewicz2022vision, chen2024catnips}, but incur the cost of volumetric rendering. Gaussian splatting replaces implicit fields with explicit primitives, enabling efficient collision evaluation. Recent methods plan on Gaussian representations through safety corridor construction \cite{chen2025splat} and differentiable collision costs \cite{andreu2025foci}. These approaches assume static scenes. We build on FOCI and evaluate route feasibility across physically modified scenes generated by hypothetical future interventions.

\textbf{World models and scene representations. }
World models provide predictive representations for planning and decision-making in embodied AI. Latent world models support reinforcement learning and long-horizon control \cite{ha2018world,hafner2025mastering,hafner2019learning}, while generative and multimodal models predict future observations, actions, and environment evolution from visual and language inputs \cite{assran2025v, bruce2024genie, driess2023palme, Brohan-RSS-23, black2026pi0visionlanguageactionflowmodel, huang2022languagemodelszeroshotplanners}. Although these approaches improve prediction and policy learning, they generally do not explicitly represent terrain mechanics, material properties, contact dynamics, or environment evolution under physical interaction. Object-centric and relational models improve interaction reasoning \cite{8461184, NIPS2016_3147da8a}, but primarily predict observations and actions rather than construct physical models for planning.

\textbf{Physics simulation for traversability and planning. }
Differentiable simulation, learned physics engines, operator learning, and physics-informed models incorporate physical structure into predictive systems \cite{10.5555/3157382.3157601, pmlr-v119-sanchez-gonzalez20a, RAISSI2019686, li2021fourier, 10.5555/3648699.3648788}. Recent physically grounded world models combine neural scene representations with explicit simulation, including Gaussian splatting coupled with continuum mechanics and MPM dynamics \cite{xie2024physgaussianphysicsintegrated3dgaussians, abou-chakra2024physically, BORYCKI2026104699}. However, planning in unstructured environments still requires reasoning about terrain properties and interaction dynamics that govern route feasibility and may remain spatially varying, partially observable, and interaction-dependent.


\section{Constructing Physically Viable Planning Environments}

We study long-horizon robot navigation under future physical changes to the environment. Given a mapped 3D scene, a start location, and a goal location, we consider a set of physically simulated environmental interventions such as flooding or terrain collapse that alter traversability. The objective is to determine whether a route planned in the original environment remains feasible after these changes occur, and if not, how route cost, reachability, and mission feasibility are affected.
Unlike planning under perception or mapping uncertainty, we assume the original map accurately reflects the current environment. The challenge is not to improve the current world estimate through additional sensing, but to evaluate how future physical changes to the environment may affect navigation before execution.
This problem is particularly important in constrained terrain where route commitment matters. Environmental changes can remove critical crossings, disconnect reachable regions, or eliminate safe return paths after a robot has already committed to a route. In these settings, reactive replanning may be insufficient because the robot may only discover that a route has become infeasible after entering terrain from which recovery is difficult or impossible.

We characterize this across intervention levels using four measures: whether the original route persists without detour, how much longer alternative routes become when rerouting is required, how the reachable area from the start contracts as the intervention partitions navigable space, and whether any feasible path to the goal exists at all.






\subsection{Scene Reconstruction and Terrain Model}
\label{sec:terrain}
 We derive a 2D digital elevation model (DEM) from the scene reconstruction, which provides the Gaussian primitives $\{(\boldsymbol{\mu}_i, \boldsymbol{\Sigma}_i, \mathbf{c}_i, \alpha_i)\}_{i=1}^N$, where $\boldsymbol{\mu}_i \in \mathbb{R}^3$ is the mean, $\boldsymbol{\Sigma}_i \in \mathbb{R}^{3 \times 3}$ the covariance, $\mathbf{c}_i \in \mathbb{R}^{3 \times 16}$ the degree-3 spherical harmonic color coefficients, and $\alpha_i \in \mathbb{R}$ the opacity. 
 Classifying Gaussians into ground, obstacle, and floater categories is non-trivial in natural outdoor scenes without semantic labels. Floaters are artifact primitives the reconstruction produced, and standard filters based on opacity or covariance are insufficient because floaters frequently exhibit high opacity and normal covariance in outdoor reconstructions (see Appendix~\ref{app:filtering}). Our filters instead exploit the physical property that ground surfaces produce flat, horizontally-oriented Gaussians while floaters and obstacles produce vertically elongated or isotropic primitives.
 Ground Gaussians are selected using two sequential filters. First, we retain only Gaussians whose vertical extent is smaller than a threshold fraction of their horizontal extent. This removes vertically elongated primitives, which typically correspond to walls, vegetation, and reconstruction artifacts, while retaining those associated with the ground; we then remove cells where the maximum $z$ value deviates from the local neighborhood median by more than $\Delta h_{\max}$, preventing floating Gaussians and aerial artifacts from entering the terrain surface. The remaining Gaussians primitives are binned into a grid at cell size $\Delta c$, and the maximum $z$ value per cell defines the DEM: $h(x,y) = \max_{i:(x_i,y_i)\in\mathrm{cell}(x,y)} \mu_{i,z}$.
%
 
The obstacle mask is computed over the same grid. A cell is marked as impassable if any Gaussian within its column exceeds the local DEM surface by more than a traversability threshold $\tau_{\mathrm{obs}}$, flagging steep terrain, boulders, and above-ground clutter. Together $h$ and the obstacle mask form a 2D binary occupancy map $\mathcal{G}_0$. This is illustrated in Figure \ref{fig:dem}.

\subsection{Intervention Simulation}
\label{sec: simulation}
We simulate environmental interventions by applying MPM-based physics to Gaussian splat scenes over surface and subsurface material points, extending methods such as PhysGaussian \cite{xie2024physgaussian} to support multi-material interactions and long-horizon rollouts. Different interventions use different constitutive models, including fluids for flooding, granular materials for landslides, and elastic models for trees and other obstacles. Each simulation modifies the baseline reconstruction by displacing, depositing, or removing Gaussian primitives, producing a physically consistent scene.

We run simulations across a range of intervention magnitudes and project each resulting scene into an occupancy grid using an intervention-specific navigation threshold. Cells become impassable when intervention-generated Gaussians exceed the threshold, corresponding to water depth for flooding and obstacle height for deposited material such as landslide debris or fallen trees.


Each MPM simulation output is projected into three derived quantities used by the planner. Let $h_{\mathrm{int}}(x,y)$ denote the intervention surface height at cell $(x,y)$, extracted from the simulation: for flooding this remains at the water level, and for deposited material such as landslide debris it is the deposited obstacle height above the terrain surface $h(x,y)$. Given a wading threshold $\tau_{\mathrm{wade}}$, we define the impassable mask: $\mathcal{M}_i(x,y) = \mathbf{1}[h_{\mathrm{int}}(x,y) - h(x,y) > \tau_{\mathrm{wade}}]$,
%
%
which flags cells where intervention depth exceeds the threshold as hard no-go regions. 
The navigation surface is defined by $h_{\mathrm{nav}}(x,y)=h_{\mathrm{int}}(x,y)$ when $\mathcal{M}_i(x,y)=0$ and the intervention deposits solid material, and $h_{\mathrm{nav}}(x,y)=h(x,y)$ otherwise. Thus, the robot rides deposited solid material where traversable and the dry terrain surface otherwise, including through shallow fluid interventions.
These quantities are produced independently for each simulation magnitude $i$ and passed to the planner in Section~\ref{sec: planning}.

\subsection{Traversability-Aware Planning}
\label{sec: planning}

We build on the Gaussian overlap collision integral and B-spline trajectory optimization framework from FOCI~\cite{andreu2025foci}, which represents both the robot and the scene as Gaussian primitives and minimizes their analytic overlap integral, solved with interior point method (IPOPT)~\cite{wachter2006implementation}. We extend this with terrain-following initialization, a heading alignment cost, per-sample ground-contact constraints, volumetric hazard encoding for intervention-affected cells, and trajectory z-pinning. Together these additions enforce a physically consistent intervention model across initialization, optimization, and postprocessing.

\paragraph{Optimization Formulation}
 
The trajectory is parameterized as an unclamped uniform cubic B-spline with $H$ control points $\mathbf{Q} = [\mathbf{q}_1, \ldots, \mathbf{q}_H]^\top \in \mathbb{R}^{H \times 4}$ over $(x, y, z, \theta)$, where $\theta$ is yaw, and discretized into $K$ equidistant samples along the progress variable $s \in [0, H{-}3)$ following a similar parameterization to that used in~\cite{andreu2025foci}. Interpolating the spline at $K$ points yields a discretized trajectory $\{ \mathbf{x}_k \}_{k=1}^{K}$ with $\mathbf{x}_k = [x_k, y_k, z_k, \theta_k]^\top$ and its time derivatives as $\dot{\mathbf{x}}_k, \ddot{\mathbf{x}}_k, \dddot{\mathbf{x}}_k$.

As in FOCI, we represent the robot as $M$ Gaussian primitives $\{(\bar{\boldsymbol{\mu}}_j, \bar{\boldsymbol{\Sigma}}_j)\}_{j=1}^M$, where $\bar{\boldsymbol{\mu}}_j(\mathbf{x}_k) \in \mathbb{R}^3$ is the mean position of robot Gaussian $j$ at state $\mathbf{x}_k$ via a three-point kinematic model (front, center, rear), and $\bar{\boldsymbol{\Sigma}}_j \in \mathbb{R}^{3\times3}$ is its covariance encoding the robot's spatial extent. We compute the collision cost as the analytic overlap integral between the robot Gaussians and the $N$ scene Gaussians $\{(\boldsymbol{\mu}_i, \boldsymbol{\Sigma}_i)\}_{i=1}^N$ from the reconstruction:
\begin{equation}
  C(\mathbf{x}_k) =
    \sum_{i=1}^{N} \sum_{j=1}^{M}
    \mathcal{N}\!\left(
      \bar{\boldsymbol{\mu}}_j(\mathbf{x}_k);\,
      \boldsymbol{\mu}_i,\,
      \boldsymbol{\Sigma}_i + \bar{\boldsymbol{\Sigma}}_j
    \right),
  \label{eq:collision}
\end{equation}
where $\mathcal{N}(\cdot) := \exp\bigl(-\tfrac{1}{2} (\bar{\boldsymbol{\mu}}_j - \boldsymbol{\mu}_i)^\top (\boldsymbol{\Sigma}_i + \bar{\boldsymbol{\Sigma}}_j)^{-1} (\bar{\boldsymbol{\mu}}_j - \boldsymbol{\mu}_i)\bigr)$ is the exponential of the negative Mahalanobis distance under the joint covariance of the scene and robot Gaussians. Because the interaction cost can be easily parallelized and GPU-accelerated, we implement it in NVIDIA Warp~\cite{Macklin_Warp_A_High-performance_2022}.

A heading alignment term $J_{\mathrm{head}} = \sum_{k=1}^{K}(\cos\theta_k \dot{y}_k - \sin\theta_k \dot{x}_k)^2$ penalizes deviation between the robot's yaw $\theta_k$ and its velocity direction $(\dot{x}_k, \dot{y}_k)$.
%
%
This term is necessary for ground robots, where yaw remains consistent with the direction of travel along terrain. We solve the following nonlinear program with IPOPT~\cite{wachter2006implementation}:
\begin{align}
  \min_{\mathbf{Q}} \quad
    & \omega_0 \sum_{k=1}^{K} \|\dddot{\mathbf{x}}_k\|^2
    + \omega_1 \sum_{k=1}^{K} C(\mathbf{x}_k)
    + \omega_2 \|\mathbf{x}_K - \mathbf{x}_{\mathrm{goal}}\|^2
    + \omega_3\, J_{\mathrm{head}}
  \label{eq:opt} \\
  \text{s.t.} \quad
    & \|\mathbf{x}_1 - \mathbf{x}_{\mathrm{start}}\|^2 \leq \epsilon,
      \notag \\
    & \|\dot{\mathbf{p}}\|^2 \leq v_{\max}^2, \quad
      \|\ddot{\mathbf{p}}\|^2 \leq a_{\max}^2,
      \quad \forall s \in [0, H{-}3), \notag \\
    & h_{\mathrm{nav}}(x_k, y_k) + c \leq z_k
      \leq h_{\mathrm{nav}}(x_k, y_k) + c + b,
      \quad \forall k,
      \notag
\end{align}
where $\mathbf{p} = [x, y, z]^\top$ denotes the position component of the state as we do not constrain the turn rate of the robot, $c$ denotes robot clearance above the navigation surface, and $b$ is the allowable vertical band. $\epsilon$ is a small start-constraint tolerance, and $\omega_0, \omega_1, \omega_2, \omega_3$ are cost weights. The navigation surface $h_{\mathrm{nav}}$ and impassable mask $\mathcal{M}_i$ are defined in Section~\ref{sec: simulation}. Velocity and acceleration bounds are enforced continuously for all $s \in [0, H{-}3)$ via convex hull constraints on the MINVO spline basis~\cite{tordesillas2022minvo}, which guarantees constraint satisfaction along the entire trajectory without requiring dense sampling. The terrain band constraint keeps each discretized state within a navigable vertical band above $h_{\mathrm{nav}}$, stabilizing optimization and preventing the optimizer from routing above the column-filled obstacle volume to avoid intervention cost.

\paragraph{Graph Search With Terrain-Based Cell Masking}
A voxel-based graph search generates a suitable initialization for the trajectory optimization step.
Standard 3D free-space search treats $z$ as a free dimension. This fails for ground robots navigating intervention-affected terrain: a 3D search often finds trajectories that exploit verticality to pass above or below a thin intervention layer, producing paths that are collision-free in the voxelized search space but physically impossible for a ground-contact vehicle. Instead, we search over a navigation surface $h_{\mathrm{nav}}$ with neighbor expansion restricted to $xy$. At each candidate cell $(x', y')$, $z' \in [h_{\mathrm{nav}}(x', y') + c, h_{\mathrm{nav}}(x', y') + c + b]$ is constrained to a local band above the surface.
%
%
Cells flagged in the impassable mask $\mathcal{M}_i$ are blocked as hard constraints and the search never expands into them. This forces lateral reasoning around impassable cells and correctly captures the mobility constraints of a ground-contact vehicle.
 
\paragraph{Volumetric Hazard Encoding}

The impassable mask $\mathcal{M}_i$ correctly blocks intervention-affected cells during search, but a spline optimizer operating on sparse scene Gaussians alone can smooth waypoints back through those cells along shorter paths. 
To account for this during optimization, we construct a vertical column of Gaussian obstacle primitives spanning the intervention volume. Let $h(x,y)$ be the dry terrain DEM and $h_{\mathrm{int}}(x,y)$  the intervention surface height from the MPM simulation output, The obstacle set is defined as
\begin{equation}
  \mathcal{O}_i =
  \bigl\{\,\boldsymbol{\mu} \in \mathbb{R}^3 :
    (x,y) \in \mathcal{M}_i,\;\;
    h(x,y) \leq \mu_z \leq h_{\mathrm{int}}(x,y)
  \,\bigr\}.
  \label{eq:column}
\end{equation}
These obstacles enter the collision cost $C(\mathbf{x}_k)$ in~\eqref{eq:collision} alongside the original scene Gaussians, making impassable cells volumetrically costly at any height inside the intervention volume. 

Importantly, after optimization, each trajectory state is projected onto the terrain surface, $z_k = h(x_k, y_k) + c$, where $h(x,y)$ is the dry terrain DEM. Since the impassable mask $\mathcal{M}_i$ prevents the optimizer from routing through deep water, all waypoints lie over traversable terrain and the projection correctly places the robot on the ground surface.

\section{Experimental Environment and Planning Setup}

A static reconstruction captures scene geometry under the conditions present during data collection and cannot represent how changing environmental conditions affect traversability. The PVWM augments the reconstruction with physical simulation, enabling planners to evaluate route feasibility under alternative conditions before deployment.

We evaluate the approach in three settings spanning route replanning, route failure, and counterfactual hazard assessment. In a large-scale field site in the Central Texas Hill Country, flooding alters traversable crossings, forcing rerouting and, in some cases, eliminating all feasible paths to the goal. In an Alaska village scene, flooding renders a crossing infeasible despite unchanged geometry, demonstrating that route feasibility depends on physical interactions absent from static reconstructions. In a controlled sandbox, we introduce simulated landslides that are not present in the captured scene and show how the planner incorporates the resulting runout regions into route generation. Together, these experiments demonstrate how physically viable world models transform static reconstructions into predictive planning environments capable of evaluating observed and hypothetical hazards. All scenes are reconstructed from real imagery and video.

\textbf{Outdoor Field Site: Central Texas Hill Country}

\begin{figure}[t]
    \centering
    \begin{minipage}{0.75\textwidth}
    \centering
    \includegraphics[width=0.47\linewidth]{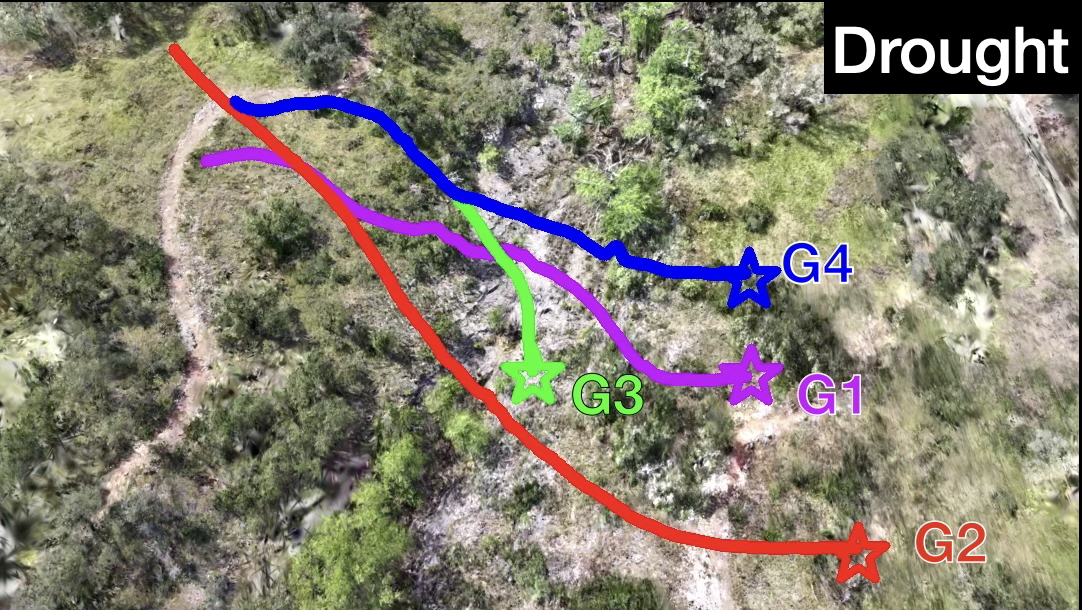}%
    \hspace{0.02\linewidth}%
    \includegraphics[width=0.47\linewidth]{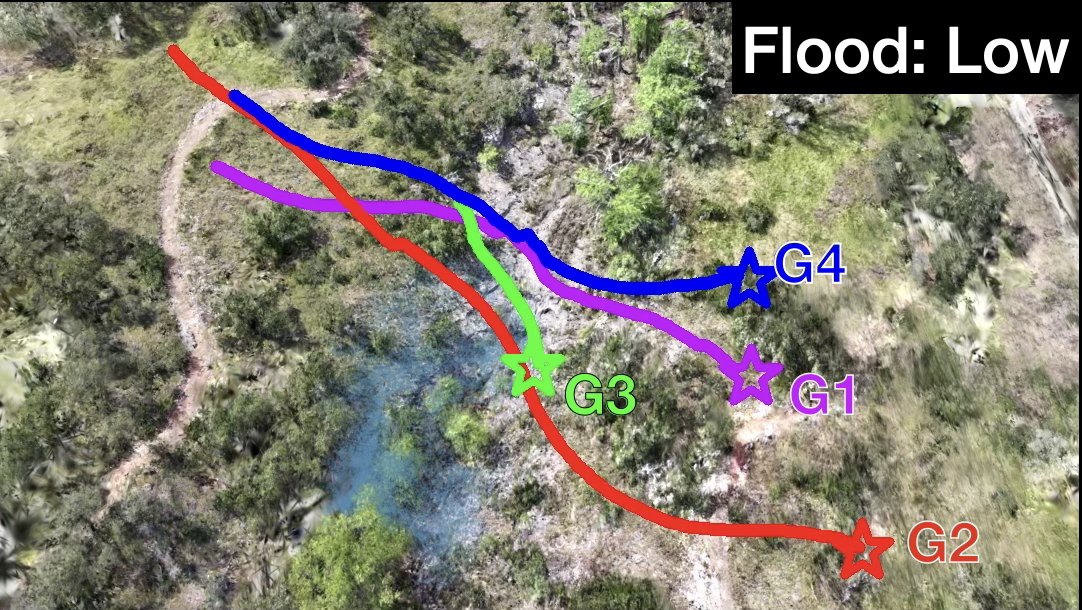}
    \vspace{-0.4em}
    \includegraphics[width=0.47\linewidth]{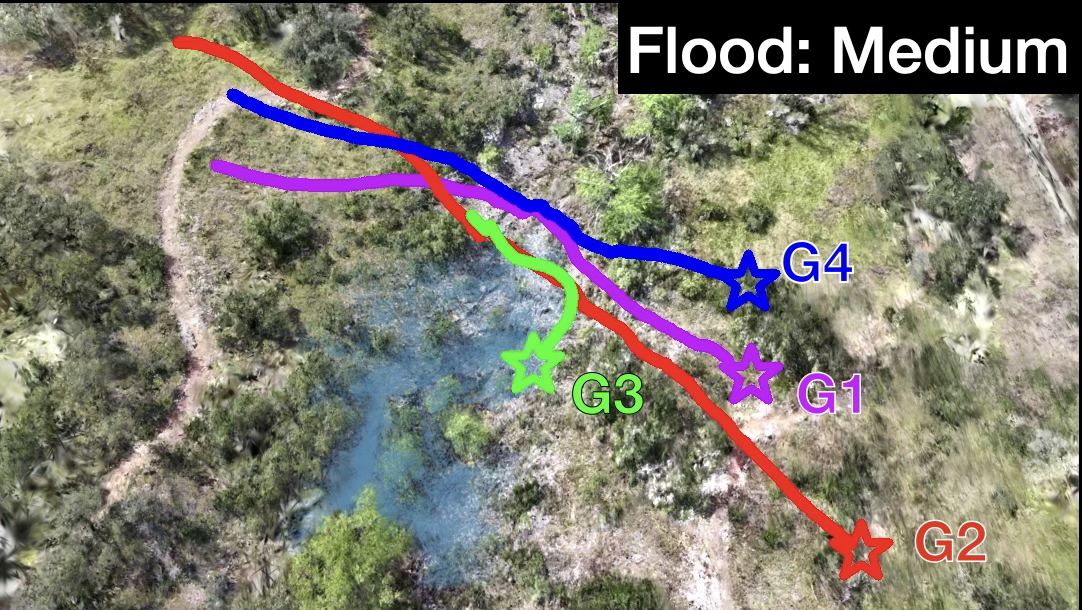}%
    \hspace{0.02\linewidth}%
    \includegraphics[width=0.47\linewidth]{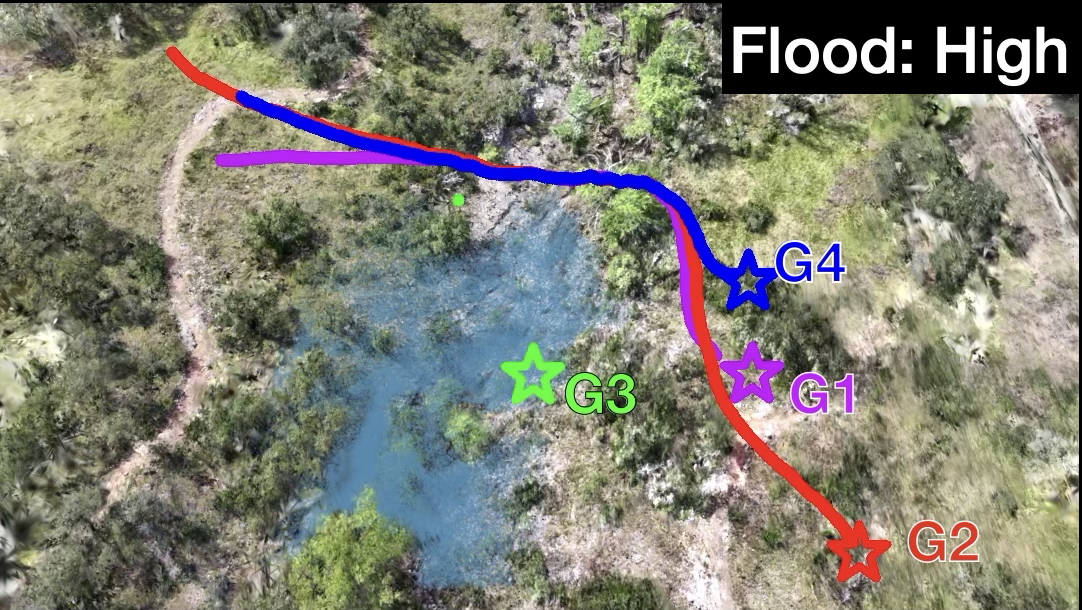}
    \end{minipage}
    \caption{Path planning results under different flood levels.}
    \label{fig:ravine}
\end{figure}

We evaluate the proposed approach on a real outdoor field site in the Central Texas Hill Country, a region characterized by steep limestone terrain, narrow drainage channels, and frequent flash flooding. The site spans approximately $500 \times 500$ ft and contains a ravine intersected by a flood-prone drainage channel. During dry conditions, a ground vehicle can traverse several crossings throughout the ravine, but rising water levels may partially or completely block those same crossings. Because steep embankments limit recovery once a vehicle enters the channel, route feasibility depends on environmental conditions that evolve after the scene is reconstructed.

We collected approximately 21 minutes of 4K aerial video using a drone platform, recovered camera poses with COLMAP, and trained a 3D Gaussian splatting reconstruction of the terrain. We then assigned physical material properties to the reconstructed surface and sampled subsurface points, producing a physically viable world model capable of simulating terrain-fluid interactions. To model changing conditions, we use the material point method to release varying water volumes into the drainage channel and ravine, generating multiple flood severities from a single drought reconstruction. Appendix~\ref{app:fieldsite} reports reconstruction quality, material assignments, and simulation parameters.

For each of four start-goal pairs, we plan over the resulting occupancy grids and report the rerouting cost ratio $\delta_i = L(\rho^*i) / L(\rho{\mathrm{drought}})$ relative to the drought route (Table~\ref{tab:reroute_ratio}, Fig.~\ref{fig:ravine}). The planner blocks cells whose flood depth exceeds the wading threshold $\tau_{\mathrm{wade}}$ while retaining shallower cells as traversable. The four pairs produce three outcome profiles that are invisible in the drought reconstruction alone. Pairs 1 and 2 remain feasible at all flood levels, with rerouting cost of Pair 1 increasing from small detours at low severity ($+0.4\%$) to substantial rerouting at high severity ($+20.3\%$). Pair 2 and 4 follows the same trend with smaller growth ($+7.2\%$ at high severity). Pair 3 incurs only minor detours at low and medium severity, yet becomes unreachable under severe flooding, rendering the mission infeasible. Because the ravine prevents safe reversal after channel entry, evaluating the physically viable world model before deployment exposes this failure mode and enables replanning before commitment. A static drought reconstruction provides no indication of any of these outcomes.

\subsection{Route Failure Under Flooding: Alaska Village Scene}

\begin{figure}[!t]
    \centering
    \includegraphics[width=\linewidth]{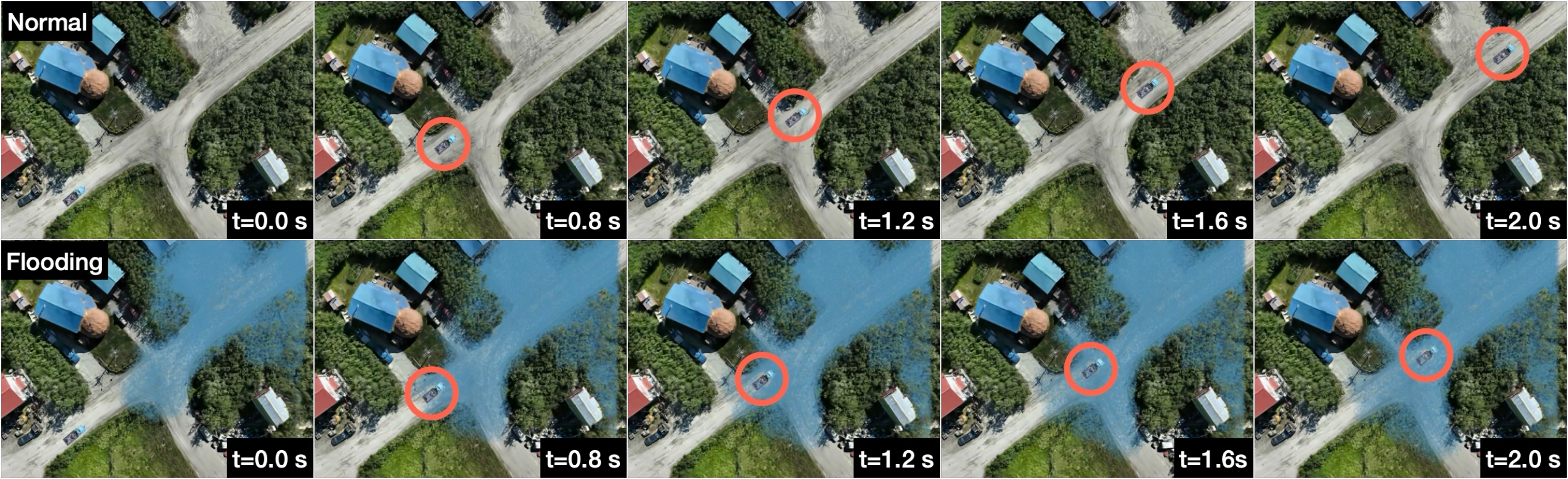}
    \caption{Simulating traversability of an Alaskan Village before and after flood. Top: the original reconstruction of the Alaska dataset with a truck simulated as a rigid ball applied to constant torque. Bottom: flood simulation using additional material points interacting with the rigid truck. The orange circles highlight the truck location in each frame.}
    \label{fig:alask}
\end{figure}

We next consider a real village environment where flooding changes route feasibility without altering the scene geometry. We reconstruct an Alaskan village from drone imagery as a 3D Gaussian splatting scene containing a road intersection and surrounding terrain, then simulate a ground vehicle traversing the intersection under dry and flooded conditions. Both scenarios use the same reconstruction; the only difference is the presence of floodwater and its interaction with the vehicle. We model the terrain as a granular material using the material point method (MPM). A truck splatting model traverses the scene while an underlying rigid body governs its motion. For the flood scenario, we introduce water as additional material points and allow it to accumulate before the simulation. Appendix~\ref{app:alaska} reports reconstruction statistics, sampling procedures, and material parameters.

Figure~\ref{fig:alask} compares the resulting trajectories. Water is rendered as semi-transparent blue splats, and orange circles indicate the truck position. Under dry conditions, the truck successfully traverses the intersection and follows the intended route. Under flooded conditions, the vehicle enters the inundated region, loses momentum through interaction with the water, and experiences a substantial reduction in mobility. As flooding increases, the vehicle reaches a critical submerged depth before completing the crossing, causing the route to transition from feasible to infeasible despite identical geometry. This outcome cannot be predicted from a static reconstruction alone because only the environmental conditions change. The physical simulation captures how those conditions alter vehicle-terrain interactions and, ultimately, route feasibility.

\subsection{Counterfactual Hazards: Toy Sandbox}

\begin{figure}[!t]
    \centering
    \includegraphics[width=\linewidth]{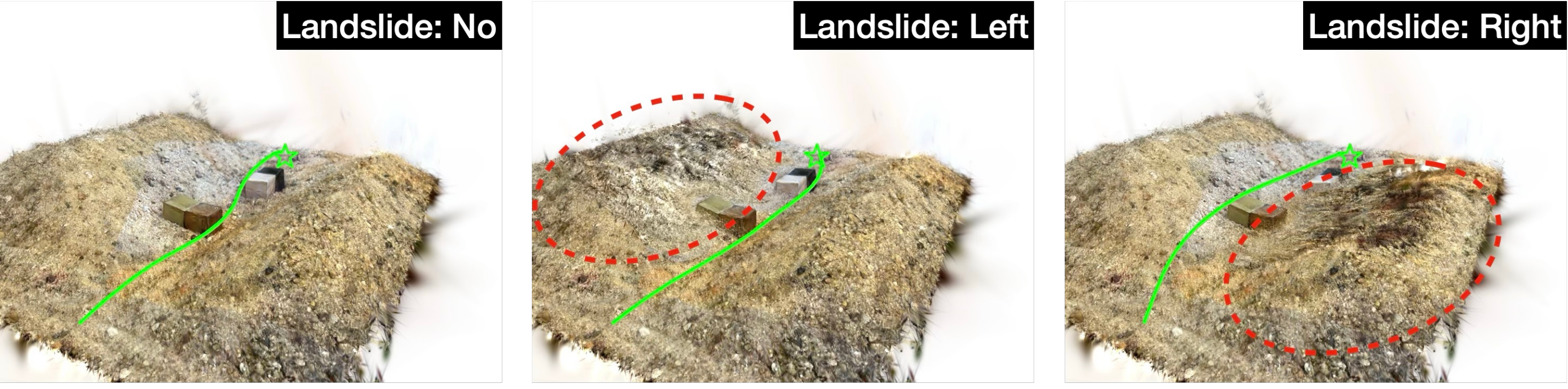}
    \caption{Path planning in the sandbox. The red dashed circle highlights the landslide region.}
    \label{fig:sandbox}
\end{figure}




The field experiments focus on hazards arising from environmental conditions present at the site. We also consider physically plausible hazards that do not exist in the captured scene, allowing the world model to evaluate hypothetical events before deployment.

To demonstrate this capability, we construct a sandbox environment consisting of a granular valley with a narrow central channel and four obstacle cubes. We reconstruct the scene as a Gaussian splatting model and simulate landslides with the material point method. Appendix~\ref{subsec:sandbox} provides reconstruction and simulation details. We evaluate planning under three conditions: no landslide, a collapse on the left side of the valley, and a collapse on the right. In the baseline scene, the planner routes through the central passage to reach the goal. In the landslide scenarios, collapsing material fills the passage and blocks the nominal route, causing the planner to incorporate the resulting runout region into the occupancy map and generate an alternative path around the obstruction (Fig.~\ref{fig:sandbox}).

\section{Conclusion \& Limitations}

We presented a physically viable world model for evaluating long-horizon navigation under future terrain change. By combining scene reconstruction, physics-based simulation, and terrain-aware route evaluation, the system measures how environmental changes affect route feasibility, reachability, rerouting cost, and mission feasibility. Our results show that physically modified environments can reveal navigation failures that are not apparent from the original reconstruction alone.

\textbf{Limitations. } 
Nevertheless, several limitations remain. Traversability is estimated primarily from geometry and does not account for semantic information or material properties, limiting both terrain understanding and automatic selection of appropriate physical models. The navigation representation is restricted to a 2D occupancy map, which cannot capture vertical structure and may produce overly conservative obstacle estimates, especially around trees and foliage. Route evaluation is performed over a discrete set of intervention levels, potentially missing important transitions in feasibility between simulated conditions. In addition, the hypothetical environments evaluated in this work exist only in simulation; while necessary for reasoning about future terrain change, the resulting assessments depend on the fidelity of the underlying physical model. This also limits the applicability of hardware validation. 

\bibliography{bibliography}  

\newpage
\appendix

\section{Additional Experiments}

\subsection{Central Texas Hill Country}
\label{app:fieldsite}

We evaluate the system on a real outdoor field site, an approximately
$500 \times 500$ ft area of steep limestone terrain in the Central Texas Hill
Country. The site contains a narrow ravine and a flood-prone drainage channel.
A ground vehicle can cross the ravine during drought, but rainfall and flash
flooding fill the channel and can remove feasible crossings, which makes
the site a demanding long-horizon planning case. The steep geometry also makes
route commitment consequential, since a robot that descends into the channel
may be unable to reverse safely once conditions change.

\textbf{Data collection and reconstruction. }
We collect roughly 21 minutes of 4K drone video covering the site from multiple
viewpoints and elevations. We extract frames at 3 FPS, downsample them to
$1920 \times 1080$ px to speed up reconstruction, and discard frames with weak
parallax such as those captured during near-pure drone rotation. This yields
500 images for training. COLMAP recovers all camera poses and approximately $43{,}000$ sparse
points. We then train a Gaussian splatting model for $30{,}000$ iterations,
producing approximately 7 million splats with a PSNR of $21.02$ and an SSIM of
$0.610$ on the validation images.

\textbf{Simulation setup. }
We crop the ravine region and rescale it to fit within a unit cube, then run MPM
on a $1/200$ grid and sample subsurface material points beneath the reconstructed
surface. We assign the surface splats a soft elastic material with a Young's
modulus of $200$ kPa and a density of $7{,}800~\mathrm{kg/m^3}$ so that surface
vegetation visibly deflects under the floodwaters, and we model the subsurface
points as granular sand using the Drucker--Prager plasticity model with a
friction angle of $45^\circ$ and a density of $1{,}600~\mathrm{kg/m^3}$. Each
simulation runs for $20{,}000$ steps at a time step of $1 \times 10^{-4}$ s.

\begin{figure*}[t]
    \centering

    \includegraphics[width=0.4\textwidth]{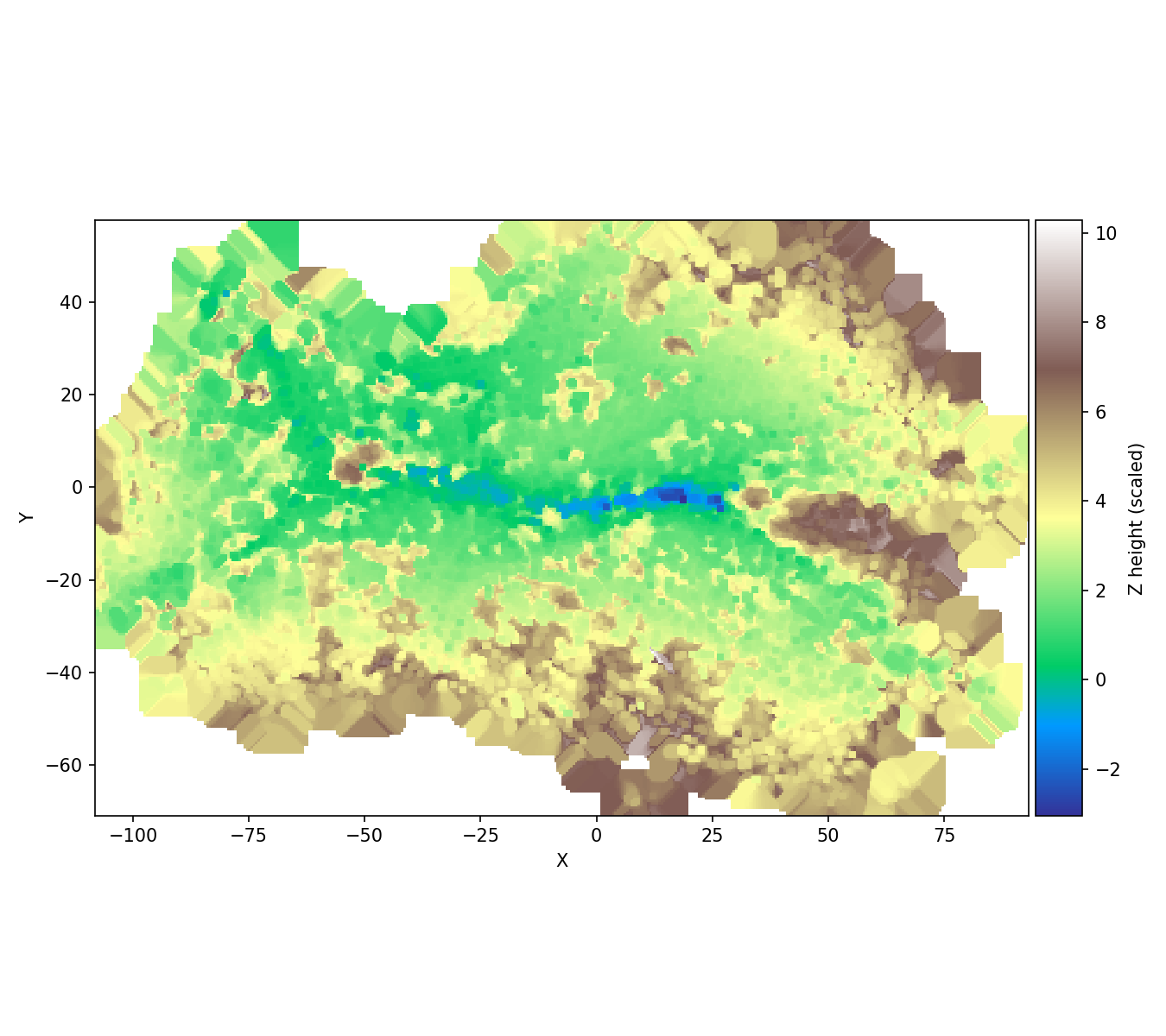}
    \includegraphics[width=0.4\textwidth]{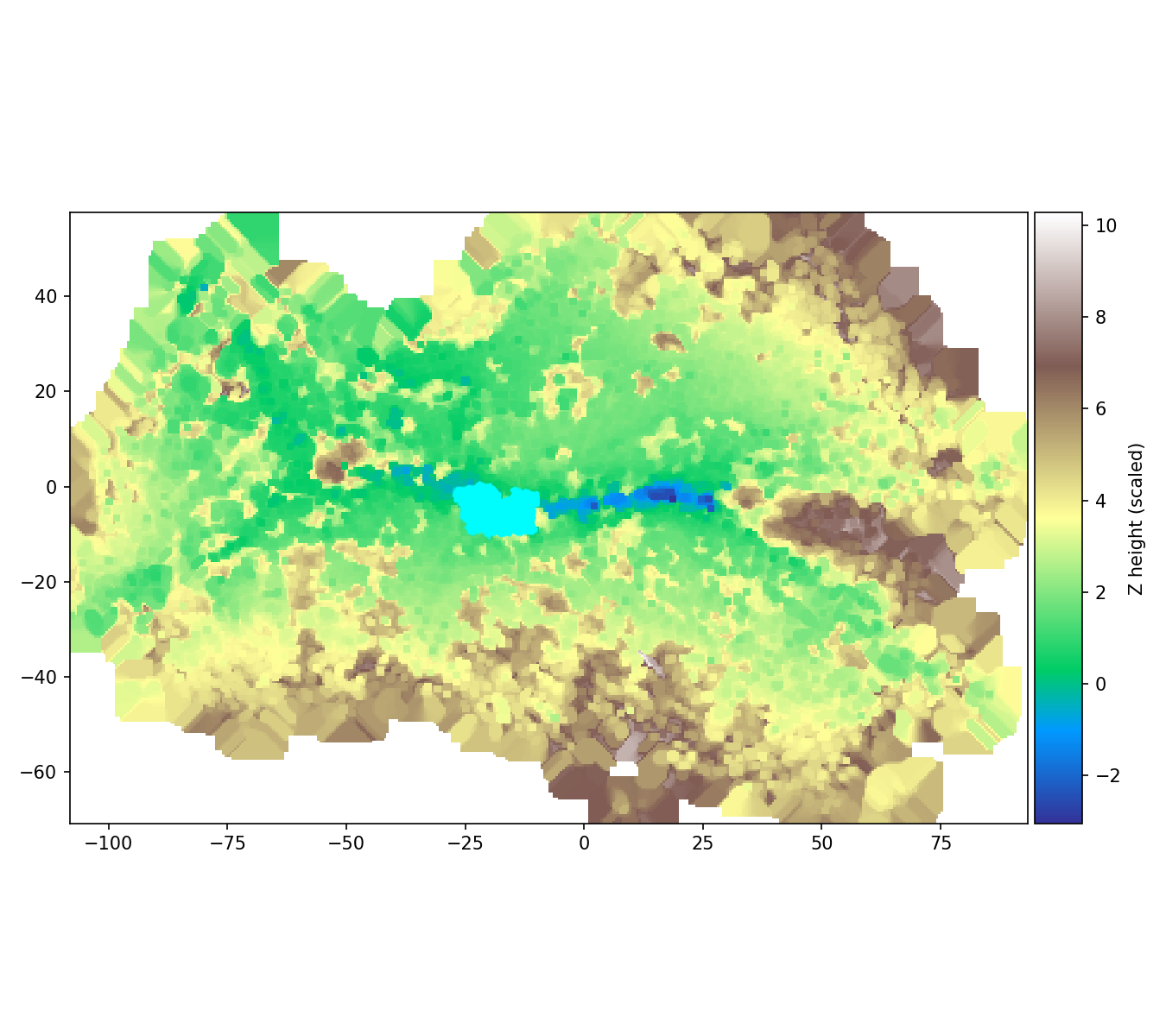}
    
    
    \includegraphics[width=0.4\textwidth]{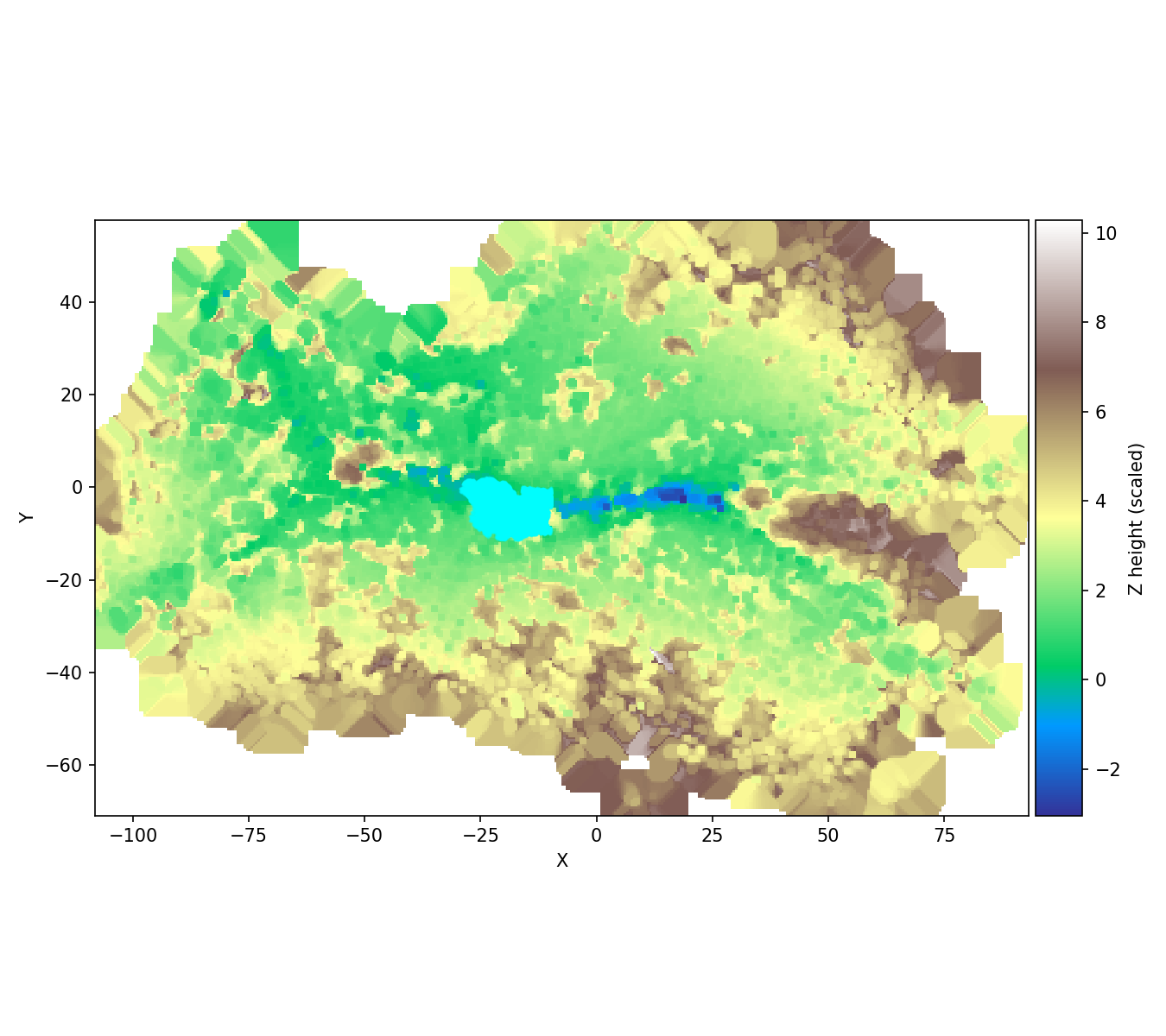}
    \includegraphics[width=0.4\textwidth]{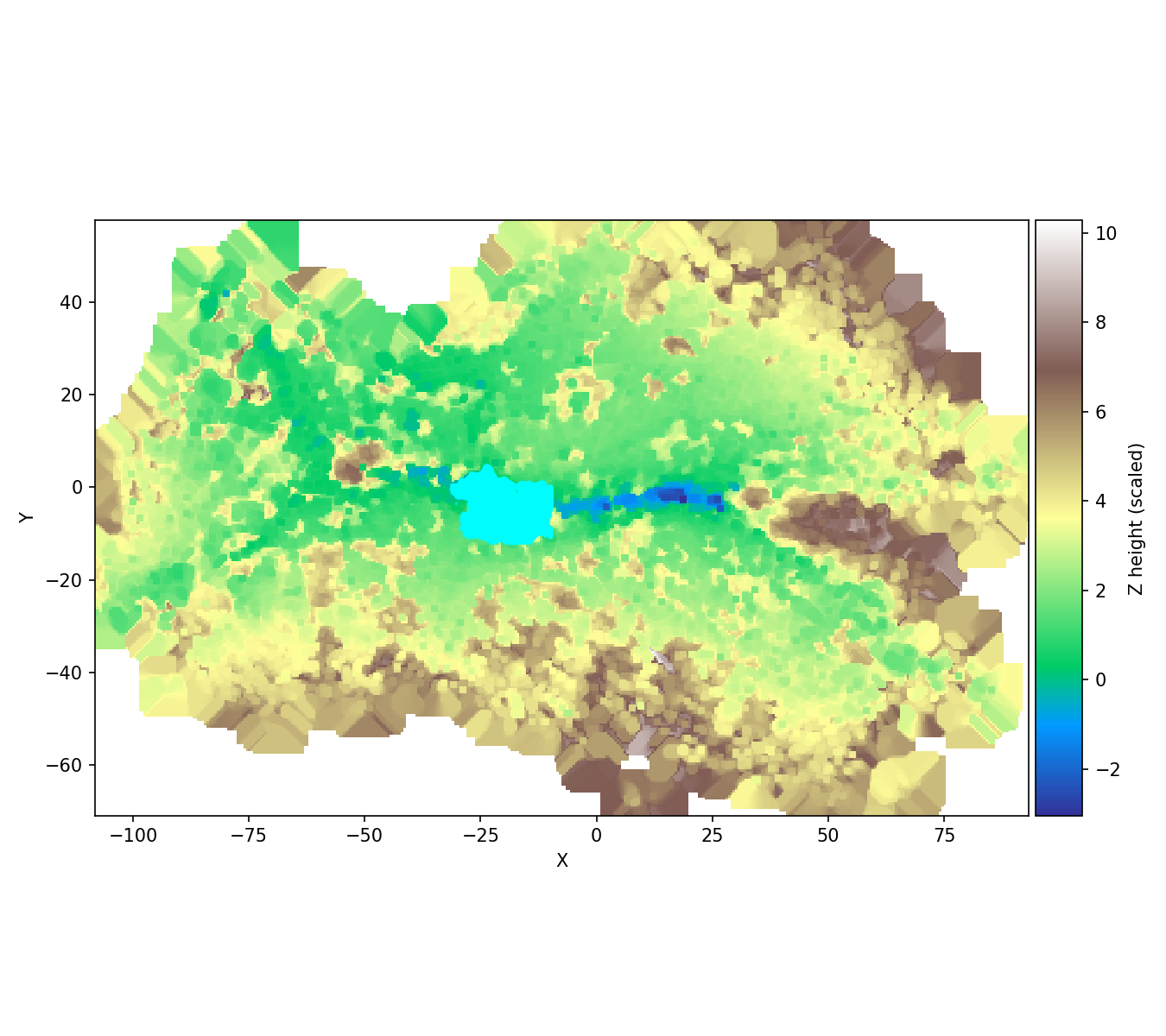}
    
    \caption{Terrain height maps under increasing flood levels. The first row corresponds to dry and low flood conditions, while the second row corresponds to medium and high flood conditions.}
    \label{fig:heightmaps}
\end{figure*}

\textbf{Flood generation. }
We simulate flooding by sampling water material points within
$0.1 \times 0.1 \times 0.05$ regions above the surface and releasing one such
water cube every 100 steps, each with an initial velocity directed toward the
ravine. We set the total number of released cubes to control flood severity:
15 cubes for low, 25 for medium, and 40 for high. Together with the drought
baseline, this gives four traversability conditions over the same scene. The height maps before and after flooding at different levels are shown in Figure \ref{fig:heightmaps}.

\begin{figure}[t]
    \centering
    \includegraphics[width=0.7\columnwidth]{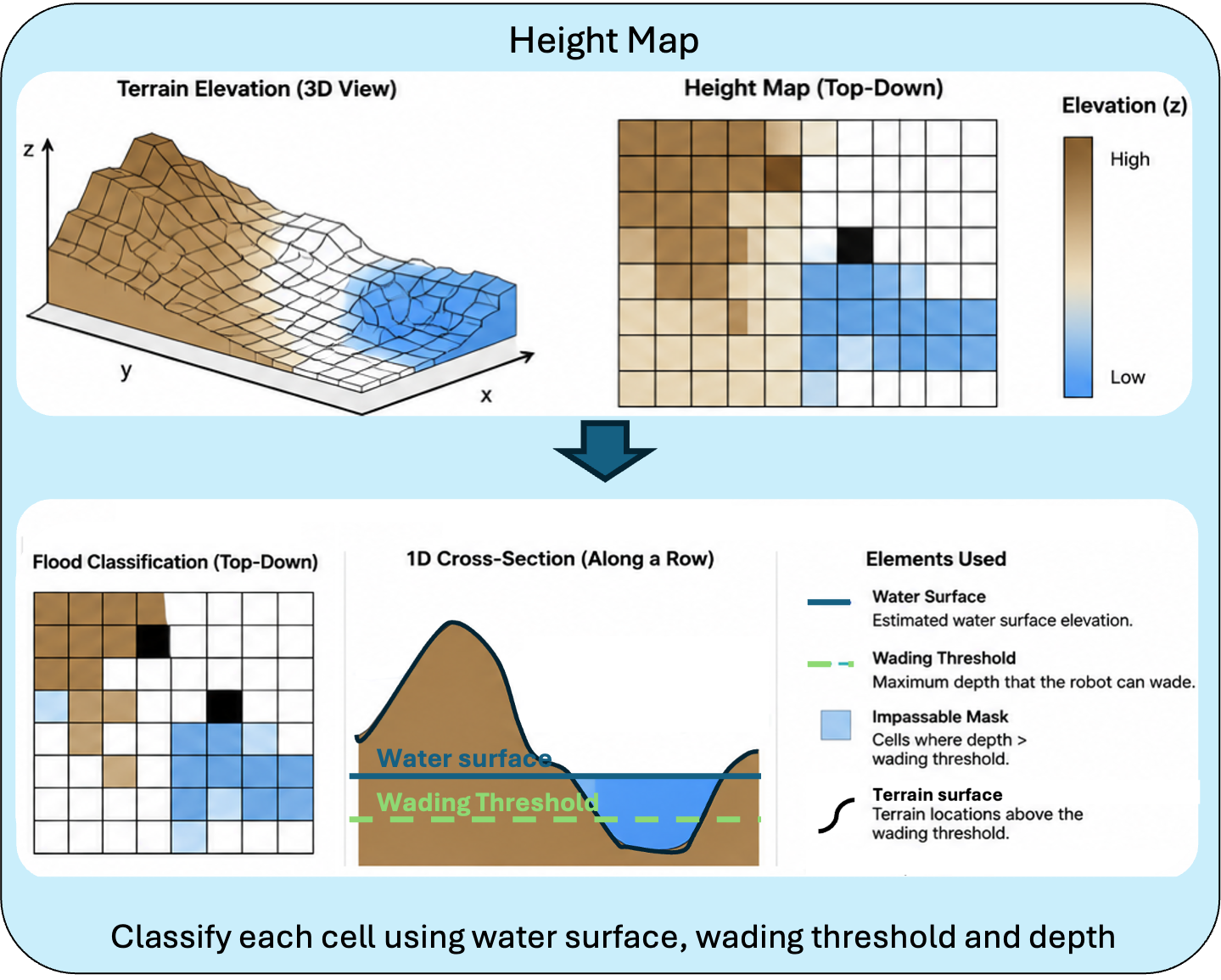}
    \caption{Digital elevation map (DEM) of the environment.}
    \label{fig:dem}
\end{figure}

\textbf{Results. }
The PVWM separates the four start-goal pairs into outcomes that the drought reconstruction cannot distinguish. Table~\ref{tab:reroute_ratio} reports the rerouting cost ratio $\delta_i = L_{\mathrm{adj}}(\rho^*_i) / L_{\mathrm{adj}}(\rho_{\mathrm{drought}})$ at each flood level, and Fig.~\ref{fig:ravine} shows the planned trajectories. A dash in Table~\ref{tab:reroute_ratio} marks an infeasible mission, where no path from start to goal exists in the modified occupancy grid $\mathcal{G}_i$.

The planner blocks only cells whose flood depth exceeds the wading threshold $\tau_{\mathrm{wade}}$ and keeps shallower cells traversable, so each reported detour reflects the planner threading a path through the shallowest available sections of a crossing. The four pairs separate by how their crossing geometry responds to increasing severity.

Pairs 1 and 2 remain mission-feasible at all levels, with rerouting cost growing progressively as flood expands: from $+0.4\%$ and $+2.3\%$ at low severity to $+20.3\%$ and $+9.7\%$ at high. Pair 4 follows a similar monotonic pattern with smaller overall growth ($1.2\%$, $+3.6\%$, $+7.2\%$ at low, medium, and high respectively), consistent with a crossing that narrows progressively without a hard threshold transition.

Pair 3 is the consequential case for mission planning. It incurs modest detours at low and medium severity ($+5.4\%$ and $+5.2\%$) but its goal region becomes unreachable at high flood, the only infeasible mission in the evaluation. Because the steep ravine prevents safe reversal once the robot commits to the channel, evaluating the physical world model before deployment surfaces this risk in advance and lets the operator abort the mission or reassign the pair. None of these outcomes is visible from the drought reconstruction alone. Evaluating the PVWM before execution lets the robot or operator stratify mission risk and choose routes that remain viable under plausible future conditions.

We provide additional details on the experiments in this section, including data collection, world model construction, simulation and more. 

\subsection{Alaska Village Scene}
\label{app:alaska}

This appendix gives the full reconstruction and simulation setup for the Alaska
village scene.

\textbf{Reconstruction. }
We reconstruct the village from high-resolution drone footage at
\(5280 \times 3956\) px. COLMAP processes 634 frames, recovers a camera pose for each,
and produces roughly \(150{,}000\) sparse points. We train the Gaussian splats for
30{,}000 iterations, which yields approximately 5 million splats. On the validation
images the reconstruction reaches a PSNR of 20.53 and an SSIM of 0.567.

\textbf{Simulation domain and terrain. }
We crop a crossroads region and rescale it to fit inside a unit cube, which defines
the MPM simulation domain. The simulation grid resolution is \(1/200\). To build the
ground, we partition the domain into a \(200 \times 200\) grid in the \(x\)-\(y\)
plane, matching the grid resolution, and sample a column of subsurface material points
under each cell. We set each column height to the 20th-percentile \(z\)-value of the
splats in that cell, a low threshold that keeps above-ground objects from inflating the
surface estimate. We assign the sampled points granular sand properties through a
Drucker--Prager plasticity model with a \(45^\circ\) friction angle and a density of
\(1{,}600~\mathrm{kg/m^3}\). The village splats remain stationary throughout.

\textbf{Truck actuation. }
We train a separate truck splatting model on the dataset
of~\citet{Knapitsch2017}. A rigid sphere sits at the center of the truck, and we drive
the truck by applying a constant torque to that sphere. The truck splats follow the
sphere's translation.

\textbf{Flood generation. }
We run two conditions, normal and flooding, on the same geometry. For the flood
condition we release 30 water cubes over the course of the simulation, one every 200
steps. We form each cube by sampling material points within a
\(0.2 \times 0.2 \times 0.002\) region above the terrain.

\textbf{Integration. }
We simulate the terrain, truck, and flood together in the MPM framework with a time
step of \(1 \times 10^{-4}~\mathrm{s}\) for 20{,}000 steps.

\subsection{Toy Sandbox Scene}
\label{subsec:sandbox}

This scene stages a hazard that the real field sites cannot produce. We build a
controlled reconstruction so that we can trigger landslides, which the
natural environments resist for two reasons: First, the real terrain is not steep
enough to drive substantial collapse. Second, where small collapses do occur, they
barely change traversability, because the robot can route over the displaced
material. The sandbox removes these limitations. It lets us demonstrate hypothetical
hazards that simulation makes available, and it lets us examine the planning
decisions the robot makes under them. The result we highlight is that the planner
avoids the collapsed channel whether we stage the landslide on the left or the
right side.

\textbf{Reconstruction. }
We build a valley-like sand terrain with elevated regions on both sides and a
narrow flat channel down the middle, then place four cubes on the valley floor as
obstacles. We capture 36 photographs with an iPhone 13 while orbiting the sandbox.
COLMAP recovers 33 camera poses and $8{,}838$ sparse points for Gaussian splatting
initialization. We train the splats for $30{,}000$ iterations, producing
approximately 2 million splats with a PSNR of $27.81$ and an SSIM of $0.796$ on the
training images. We train on all images because the imagery is small.

\textbf{Simulation setup. }
We crop and center the splats in the $xy$-plane to fit within a unit cube, run MPM
on a $1/100$ grid, and sample material points beneath the cropped surface. We
segment the cube splats by hand and assign them an elastic material with a Young's
modulus of $1$ MPa and a density of $7{,}800~\mathrm{kg/m^3}$. We simulate the
remaining splats and the sampled subsurface points as granular sand using the
Drucker--Prager plasticity model with a friction angle of $15^\circ$ and a density
of $1{,}600~\mathrm{kg/m^3}$. We apply frictionless slip boundary conditions on the
four side walls and a sticky condition at the base of the sampled points. The
simulation uses gravity of $9.81~\mathrm{m/s^2}$ in the $-z$ direction and runs for
$20{,}000$ steps at a time step of $1 \times 10^{-4}$ s.

\textbf{Landslide scenarios. }
We generate three scenarios over this setup: no landslide, a landslide on the left
side, and a landslide on the right side. The no-landslide case is the initial
simulation state. We produce a one-sided landslide by holding the opposite region
stationary, zeroing its grid momentum so that only the chosen side collapses into
the channel.

\textbf{Results. }
We define a start point and a goal point on opposite sides of the valley floor and
plan across all three scenarios. Figure~\ref{fig:sandbox} shows the planned paths
over the rendered splats. In the static scene the planner threads between the
obstacle cubes in the middle of the valley and reaches the goal. In each landslide
scene the runout from the collapsing side covers the original passage, marked by
the red dashed circle, and the planner routes around both the cubes and the runout
to reach the goal. Staging the collapse on the left and then on the right shows
that the planner avoids whichever section the landslide buries rather than
attempting to cross it.


 
 
 
 

\begin{table}[t]
\centering
\caption{Rerouting cost ratio relative to the dry baseline.}
\label{tab:reroute_ratio}
\begin{tabular}{c|c c c c}
\hline
Pair & Drought & Low & Medium & High \\
\hline
1 & 1.000 & 1.004 & 1.079 & 1.203 \\
2 & 1.000 & 1.023 & 1.031 & 1.097 \\
3 & 1.000 & 1.054 & 1.052 & -- \\
4 & 1.000 & 1.012 & 1.036 & 1.072 \\
\hline
\end{tabular}
\end{table}

\section{Physically Accurate Simulation of Gaussian-Splatted Worlds}

In this section, we explain why we choose the Material Point Method (MPM) as the physics engine for our Gaussian-splatted world model.

\begin{figure}[t]
    \centering
    \includegraphics[width=0.8\linewidth]{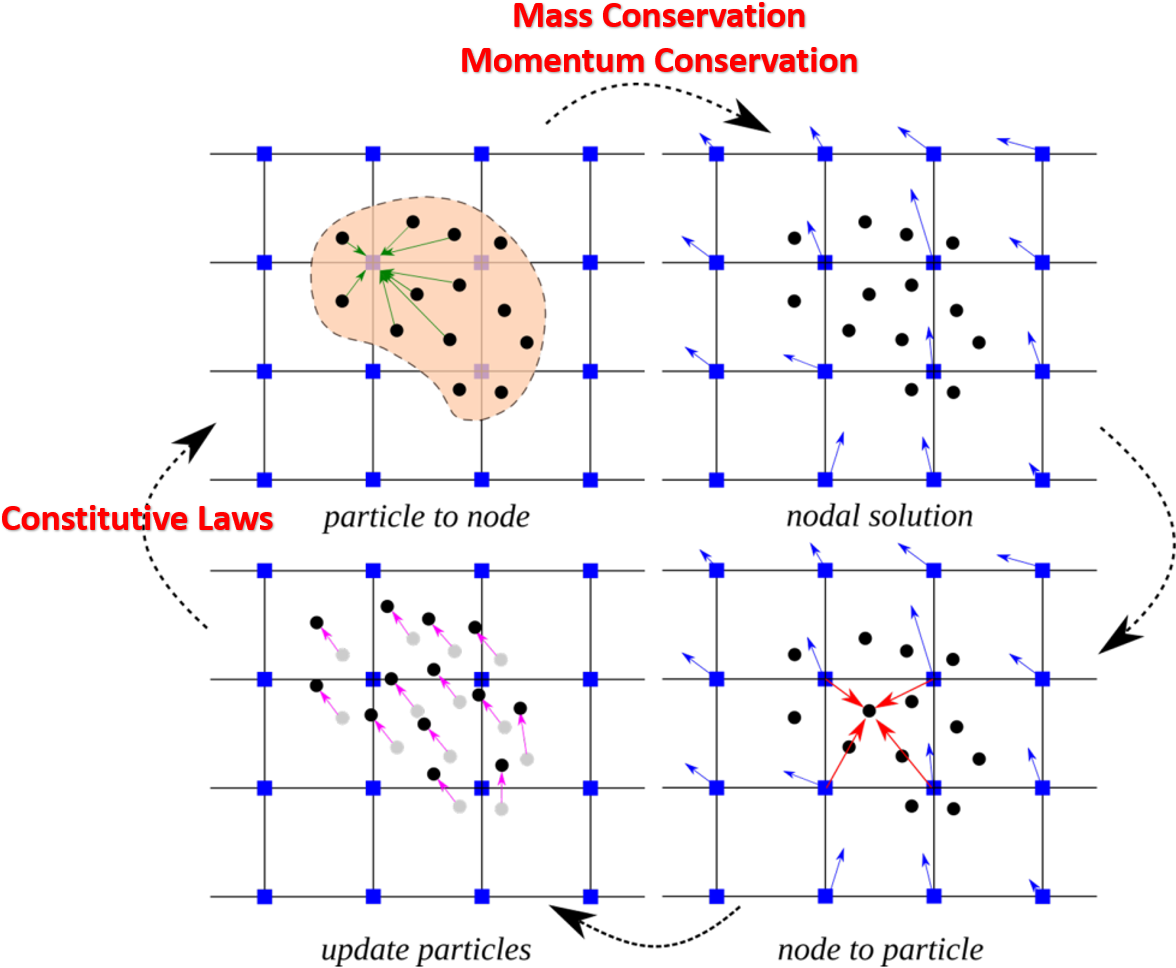}
    \caption{Illustration of the MPM update scheme. (1) Material points carry physical state (mass, momentum, stress) and are overlaid on a fixed background grid. Interpolation weights distribute each particle's mass and momentum to neighboring grid nodes (P2G). (2) The equations of motion are solved on the grid: nodal velocities are obtained by normalizing momentum by mass, internal stress forces are applied, and boundary conditions are enforced. (3) Updated nodal velocities are interpolated backe to the material points (G2P), updating each particle's velocity, position, and deformation gradient. (4) The material points advance to their new positions, and the grid is reset to zero in preparation for the next time step.}
    \label{fig:mpm_diagram}
\end{figure}

MPM is a continuum-based physics simulation method that is particularly well suited for problems involving large deformations. MPM consists of two primary components: particles and a background grid. Particles, also known as material points, represent discrete portions of a material and carry all persistent physical states, including position, velocity, mass, and deformation history. Unlike the discrete element method, interactions between particles in MPM are indirect but computationally efficient. We diagram detailing the mechanism of MPM can be found in Fig.~\ref{fig:mpm_diagram}.

During each simulation step, particles transfer information such as mass, momentum, and internal forces (or stresses) to neighboring grid nodes. The governing equations of motion are then solved on the grid. After the grid update, the resulting information is transferred back to the particles to update their velocities and positions. Collision handling occurs naturally on the grid through the accumulation and exchange of momentum among neighboring particles.

To simulate different materials, an appropriate constitutive model must be selected. The constitutive model defines the stress--strain relationship of a material and is used to compute the internal stresses and forces based on the current deformation state of each particle.

A key motivation for using MPM with Gaussian-splatted worlds is that the two representations share similar geometric structures. Each Gaussian splat is parameterized by a center position \(\mu\) and a covariance matrix \(\Sigma\), which describes its shape and orientation. These parameters naturally correspond to the position of an MPM particle, and the covariance matrix provides information about the particle's reference volume and initial deformation state. This structural correspondence between Gaussian splats and MPM particles enables direct simulation of reconstructed scenes without requiring manual mesh generation or additional geometric processing.

During simulation, the MPM particles deform and advect according to the governing physical laws. The resulting particle positions and deformation gradients are then propagated back to the corresponding splats, ensuring that the rendered scene remains consistent with the underlying physical state throughout the simulation.

The coupling between perception and physics enables the robot to reason about how a reconstructed scene will evolve under real world scenario.

\section{Obstacles Classification}
\label{app:filtering}
A fundamental challenge in outdoor Gaussian splat reconstructions is classifying Gaussians into three categories without semantic labels: traversable ground, impassable obstacles such as rocks and vegetation, and floaters, which are semi-transparent primitives the reconstruction places above or between surfaces to explain visual observations such as atmospheric haze, vegetation canopy, and view-dependent lighting effects. This classification is necessary to construct a meaningful occupancy grid: ground Gaussians define the DEM surface, obstacle Gaussians mark impassable cells, and floaters must be excluded entirely. Without semantic understanding of the scene, the classification must rely on geometric properties alone.

Floaters are the primary complication. They occupy three-dimensional space like real geometry and corrupt both the DEM and the obstacle mask if included: floaters above the ground plane appear as phantom obstacles, while floaters near the terrain surface inflate height estimates and distort the navigable surface. Common strategies for exclusion include filtering by opacity, discarding Gaussians unlikely to represent solid geometry, and filtering by minimum covariance eigenvalue, discarding small isotropic Gaussians that may represent reconstruction noise. Neither is sufficient in natural outdoor scenes: floaters frequently exhibit high opacity because the reconstruction requires them to explain observed color values, and their covariance does not reliably distinguish them from small legitimate features such as pebbles or surface texture.

We instead exploit a physical property of natural terrain to separate the three categories geometrically. Ground surfaces produce flat, horizontally-oriented Gaussians because the reconstruction fits elongated ellipsoids along the local surface plane. Obstacles such as rocks, tree trunks, and walls produce vertically-oriented or compact Gaussians. Floaters produce either vertically elongated or roughly isotropic primitives. Retaining Gaussians whose vertical extent is below a threshold fraction of their horizontal extent selects for ground-like geometry and rejects both obstacle and floater primitives. Remaining Gaussians whose maximum zz
z value deviates from the local neighborhood median by more than $\Delta h_{\max}$ are then removed as height-discontinuous artifacts. The threshold $\Delta h_{\max}$ is set empirically above the maximum expected inter-cell terrain variation: since the navigable surface across the field site spans approximately 2–2.5 m in elevation, any larger deviation identifies a floater rather than terrain.

\end{document}